\documentclass[10pt,twocolumn,letterpaper]{article}

\usepackage{cvpr}              
\definecolor{cvprblue}{rgb}{0.21,0.49,0.74}
\usepackage[pagebackref,breaklinks,colorlinks,allcolors=cvprblue]{hyperref}

\def\paperID{*****} 
\def\confName{CVPR}
\def\confYear{2026}

\title{Beyond Shortcuts: Mitigating Visual Illusions in Frozen VLMs \\ via Qualitative Reasoning}

\author{
    Hao Guo$^{1,2}$ \quad Fei Wang$^{1,3}$ \quad Junjie Chen$^{1,3}$ \quad Yiqi Nie$^{1,4}$ \quad Jiaqi Zhao$^{1,4}$ \\ \quad Qiankun Li$^5$  \quad Subin Huang$^2$ \\
    $^1$Institute of Artificial Intelligence, Hefei Comprehensive National Science Center, Hefei, China \\
    $^2$Anhui Polytechnic University, Wuhu, China \quad 
    $^3$Hefei University of Technology, Hefei, China \\
    $^4$Anhui University, Hefei, China \quad 
    $^5$IGS, Imperial College London \\
    {\tt\small \{sifan10077,ifei17.hfut,nieyiqi5,zzzendurance\}@gmail.com, q.li2@imperial.ac.uk} \\
    {\tt\small jjchen@iai.ustc.edu.cn, subinhuang@ahpu.edu.cn}
}
\begin{document}
\maketitle
\begin{abstract}
While Vision-Language Models (VLMs) have achieved state-of-the-art performance in general visual tasks, their perceptual robustness remains remarkably brittle when confronted with optical illusions. These failures are often attributed to shortcut heuristics, where models prioritize linguistic priors and memorized prototypes over direct visual evidence. In this work, we propose Structured Qualitative Inference (SQI), a training-free, data-centric framework designed to fortify visual grounding in frozen VLMs.
SQI addresses perceptual anomalies through three systematic modules: (1) Axiomatic Constraint Injection, which suppresses erroneous metric estimations and quantitative hallucinations; (2) Hierarchical Scene Decomposition, which decouples target visual manifolds from complex background distractors; and (3) Counterfactual Self-Verification, an adversarial reasoning step that mitigates confirmation bias. By orchestrating these qualitative constraints at inference time, SQI effectively aligns high-level linguistic reasoning with low-level visual perception.
Our framework was evaluated on the DataCV 2026 Challenge (Task I: Classic Illusion Understanding), where it ranked 2nd place overall. Experimental results demonstrate that SQI not only significantly enhances accuracy across diverse illusion categories but also provides superior diagnostic interpretability without any model fine-tuning. Our success underscores the potential of structured qualitative grounding as a robust paradigm for developing next-generation, illusion-resistant vision-language systems.
\end{abstract}
\section{Introduction}

Optical illusions expose a fundamental challenge for vision-language models (VLMs): distinguishing between visual appearance and underlying reality. Unlike standard visual inputs, illusionary stimuli are deliberately constructed to induce systematic perceptual errors~\cite{gregory1997knowledge}, making them a stringent test of whether models truly ground their reasoning in visual evidence. Despite their strong performance on conventional benchmarks~\cite{radford-2021-learning,li-2023-evaluating,chen2025seeingsarcasmdifferenteyes}, modern VLMs often produce confident yet incorrect judgments when confronted with such deceptive visual patterns~\cite{gomez2019convolutional}.

These failures are not due to a lack of visual information, but rather arise from how models reason about it. In particular, VLMs exhibit a strong reliance on shortcut heuristics, where linguistic priors and memorized visual prototypes override direct visual evidence~\cite{agrawal-2018-don,wang2024eulermormer,wang2024frequency}. As a result, they suffer from several characteristic failure modes, including \textit{metric hallucination}, where unreliable quantitative estimation leads to incorrect conclusions~\cite{luo2023valley}; \textit{background interference}, where irrelevant contextual patterns bias perception; and \textit{confirmation bias}, where the model over-commits to an initial hypothesis without sufficient verification~\cite{yao2022react}. Together, these issues reveal a fundamental mismatch between high-level reasoning and low-level perception.

Existing approaches to improving robustness typically focus on enhancing visual representations through data augmentation or model fine-tuning. However, such methods are often costly, task-specific, and incompatible with frozen VLMs that are increasingly used as general-purpose backbones~\cite{zhu2023minigpt}. More importantly, they do not address the root cause of the problem: the reasoning process itself.

We argue that improving robustness requires a shift from unreliable quantitative estimation to structured qualitative reasoning. To this end, we propose \textbf{Structured Qualitative Inference (SQI)}, a training-free framework that restructures inference-time reasoning in frozen VLMs. SQI enforces a set of qualitative constraints that guide models to prioritize local visual evidence over global appearance. Concretely, it consists of three components: (1) Axiomatic Constraint Injection, which suppresses unreliable metric-based reasoning~\cite{wei-2022-chain,wang2026xinsight}; (2) Hierarchical Scene Decomposition, which isolates target objects from distracting context~\cite{gu2021open}; and (3) Counterfactual Self-Verification, which mitigates confirmation bias through adversarial reasoning~\cite{madaan2023self}.

We evaluate SQI on a challenging illusion understanding benchmark, where it achieves \textbf{2nd place} in the DataCV 2026 Challenge (Task I: Classic Illusion Understanding)~\cite{illusionvqa2026}. In particular, SQI demonstrates strong robustness on perturbed inputs, highlighting its ability to resist spurious visual cues without any model fine-tuning. These results suggest that structured qualitative reasoning provides a simple yet effective paradigm for improving perceptual robustness in vision-language systems.
\section{Method}
\begin{figure*}
    \centering
    \includegraphics[width=1\linewidth]{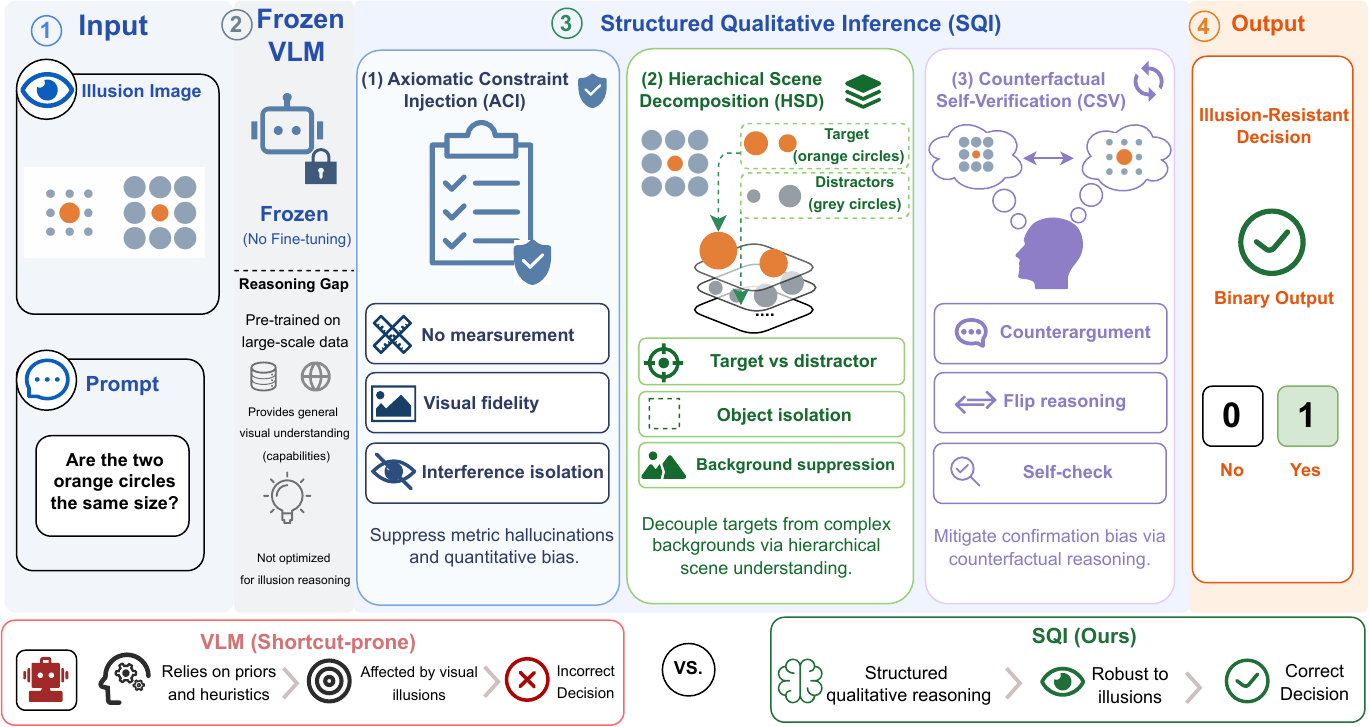}
    \caption{Overview of Structured Qualitative Inference (SQI). Given an input image and prompt, SQI enhances frozen vision-language models through three qualitative reasoning modules: (1) Axiomatic Constraint Injection to suppress metric hallucinations, (2) Hierarchical Scene Decomposition to isolate target objects from distractors, and (3) Counterfactual Self-Verification to mitigate confirmation bias. SQI significantly improves robustness to visual illusions without any model fine-tuning.}
    \label{fig:model}
\end{figure*}
\subsection{Overview}
Vision-language models (VLMs)~\cite{nie2026mer,wang2026multimodal} remain vulnerable to visual illusions due to shortcut-driven reasoning, where linguistic priors override direct visual evidence. To address this, we propose \textbf{Structured Qualitative Inference (SQI)}, a training-free framework that improves perceptual robustness by restructuring inference-time reasoning.

SQI enforces a sequence of qualitative constraints as illustrated in Figure~\ref{fig:model}. It first suppresses metric-based reasoning via \textit{Axiomatic Constraint Injection}, then isolates relevant visual entities through \textit{Hierarchical Scene Decomposition}, and finally applies \textit{Counterfactual Self-Verification} to reduce confirmation bias. 
This structured pipeline shifts the model from heuristic estimation to grounded qualitative reasoning, thereby improving robustness to illusion-induced errors.

\subsection{Formalization and Problem Statement}

We define a frozen vision-language model (VLM) as a function $\mathcal{M}(I, Q) \rightarrow A$, where $I$ denotes an image and $Q$ is a textual query about its underlying visual properties. In illusion-heavy settings, the goal is to recover the true perceptual judgment rather than the misleading appearance induced by visual patterns.
However, such models often fail due to three systemic factors: (i) \textit{metric hallucination}, where the model relies on unreliable implicit quantitative estimation; (ii) \textit{background interference}, where irrelevant visual patterns bias the perception of target objects; and (iii) \textit{confirmation bias}, where the model over-commits to its initial hypothesis driven by linguistic priors.

To address these issues, we propose Structured Qualitative Inference (SQI), formulated as an inference-time transformation:
\begin{equation}
A = \text{SQI}(\mathcal{M}, I, Q, \mathcal{C}),
\end{equation}
where $\mathcal{C}$ denotes a set of structured qualitative constraints that guide the reasoning process. Rather than directly querying the model, SQI decomposes inference into a sequence of constrained qualitative steps that suppress spurious cues and prioritize local visual evidence. This formulation operates without any model fine-tuning and serves as a lightweight reasoning-layer augmentation.

\subsection{Structured Qualitative Inference}
SQI restructures the inference process of a frozen VLM by transforming the original query into a structured qualitative reasoning protocol. The model is guided by a set of constraints $\mathcal{C}$ that enforce non-quantitative judgment, explicit scene decomposition, and self-verification within a single-pass inference. These constraints suppress spurious cues and progressively refine the model’s reasoning, without requiring any model fine-tuning.

\paragraph{Axiomatic Constraint Injection}
We impose a set of qualitative axioms that explicitly prohibit metric-based reasoning. Specifically, the model is discouraged from estimating lengths, angles, counts, or other quantitative attributes that are unreliable under illusion-induced distortions. Instead, it is guided to rely on qualitative comparison and relative visual appearance when forming judgments. This constraint shifts the model away from shortcut-driven estimation and suppresses metric hallucination.

\paragraph{Hierarchical Scene Decomposition}
We decompose the scene into target objects and background elements. The model is guided to focus on the relevant targets while ignoring surrounding distractors, such as grids, shading, or contextual patterns, that often bias perception in illusions. This encourages localized visual grounding and reduces the influence of misleading context, thereby mitigating background interference.

\paragraph{Counterfactual Self-Verification}
We re-evaluate the model’s initial judgment through counterfactual reasoning. The model is guided to consider alternative interpretations and question its initial assumption, reducing over-commitment to early predictions. This enforces consistency in reasoning and improves decision reliability, thereby mitigating confirmation bias.

\subsection{Domain-Specific Heuristic Dispatching}
Different illusion tasks exhibit distinct patterns of perceptual failure. To address this, we introduce a lightweight heuristic dispatching mechanism that adapts qualitative constraints to the query type. For example, alignment queries emphasize directional consistency while suppressing background grids, whereas color queries enforce strict isolation of target surfaces for comparison. These heuristics remain minimal and do not introduce task-specific engineering, preserving the generality of SQI.

\section{Experiment}

\begin{table}[htbp]
    \centering
    \caption{
    Performance on DataCV Challenge 2026 (Task I). 
    ``Overall'' denotes the mean accuracy over two subsets defined by ground-truth labels, where ``Pert.'' corresponds to perturbed images (GT=0) and ``Orig.'' corresponds to original images (GT=1).
}
  \label{tab:participant_acc_percent}
  \setlength{\tabcolsep}{4pt} 
  \begin{tabular}{clccc}
    \toprule
    Rank & Team name & Overall & Pert. & Orig. \\
    \midrule
    1 & snowpine007 & 71.67 & 61.90 & 81.43 \\
    \textbf{2} & \textbf{sifan077 (Ours)} & \textbf{69.05} & \textbf{67.62} & \textbf{70.48} \\
    3 & pepsí & 66.43 & 51.43 & 81.43 \\
    4 & yiming\_zhang & 65.48 & 67.62 & 63.33 \\
    5 & charles\_chen & 64.52 & 59.05 & 70.00 \\
    6 & ziq\_y & 60.00 & 60.95 & 59.05 \\
    7 & neeo333 & 59.05 & 34.76 & 83.33 \\
    8 & goudan & 58.57 & 42.38 & 74.76 \\
    9 & kishores15 & 57.14 & 59.52 & 54.76 \\
    \bottomrule
  \end{tabular}
\end{table}

\subsection{Experiment Setup}
\paragraph{Dataset.} We evaluate our method on the Task I: Classic Illusion Understanding benchmark from the DataCV 2026 Challenge~\cite{illusionvqa2026}. The dataset consists of diverse optical illusion cases where visual appearance conflicts with underlying geometric or semantic reality.

\paragraph{Evaluation Metrics.}
We follow the official evaluation protocol of the DataCV 2026 Challenge. The evaluation consists of two subsets: original images and perturbed images. We compute accuracy on each subset independently, and report the final Overall Accuracy as their average, ensuring balanced evaluation across different visual conditions.

\subsection{Main Results}
Table~\ref{tab:participant_acc_percent} reports the performance on the DataCV 2026 Challenge (Task I: Classic Illusion Understanding). Our method achieves 2nd place overall among all participating teams, demonstrating strong competitiveness under the official evaluation protocol.

Notably, SQI achieves a balanced performance across both original and perturbed subsets, with comparable accuracy on perturbed images (67.62\%) and original images (70.48\%). This indicates that our method maintains robustness under visual perturbations while preserving performance on standard inputs.

Compared to other top-ranked methods, SQI exhibits a more stable performance distribution across different visual conditions, highlighting the effectiveness of structured qualitative reasoning in improving perceptual consistency.

\subsection{Case Study}
\begin{figure}[htbp]
    \centering
    \includegraphics[width=1\linewidth]{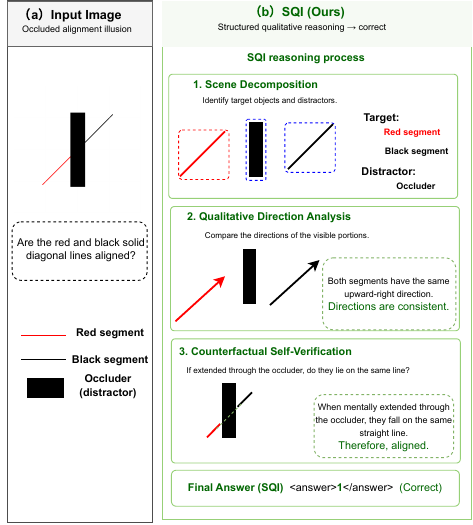}
    \caption{An example of Structured Qualitative Inference (SQI) applied to an occluded alignment illusion.}
    \label{fig:case-study}
\end{figure}
This case study exemplifies a prevalent perceptual trap where a salient occluder induces a false perception of misalignment. Standard VLMs typically default to global appearance-based shortcuts, falling prey to background interference and heuristic-driven estimations of continuity.

SQI addresses this by restructuring the reasoning process. Scene decomposition separates the segments from the occluder, qualitative direction analysis verifies consistent orientation, and counterfactual self-verification extends the segments to confirm alignment. This suppresses misleading cues and improves robustness in illusion scenarios.
\section{Conclusion}
We propose Structured Qualitative Inference (SQI), a training-free framework that improves the robustness of VLMs by restructuring inference-time reasoning. By enforcing qualitative constraints, SQI mitigates metric hallucination, background interference, and confirmation bias. Our method achieves 2nd place in the DataCV 2026 Challenge, demonstrating the effectiveness of reasoning-level interventions for illusion understanding. We hope this work inspires future research on more general and scalable inference-time alignment strategies for vision-language models.

{
    \small
    \bibliographystyle{ieeenat_fullname}
    \bibliography{main}
}


\end{document}